\documentclass[letterpaper, 10 pt, conference]{ieeeconf}
\IEEEoverridecommandlockouts
\usepackage{cite}
\usepackage{amsmath,amssymb,amsfonts}
\usepackage[english]{babel}
\usepackage{graphicx}
\usepackage{subcaption}
\usepackage{textcomp}
\usepackage[dvipsnames]{xcolor}
\usepackage[hyphens]{url}
\usepackage{hyperref}
\usepackage[normalem]{ulem}
\usepackage{comment}
\usepackage{algorithm,algorithmic}

\hypersetup{colorlinks,linkcolor={blue},citecolor={blue},urlcolor={blue}}

\begin{document}

\title{\Large{\textbf{CHARTOPOLIS: A Small-Scale Labor-art-ory for Research and Reflection on Autonomous Vehicles, Human--Robot Interaction, and Sociotechnical Imaginaries}}
}

\author{Sangeet Sankaramangalam Ulhas$^{a}$, Aditya Ravichander$^{a}$, Kathryn A.~Johnson$^{b}$, Theodore P.~Pavlic$^{c}$, \\ Lance Gharavi$^{d}$, and Spring Berman$^{a}$
\thanks{This work was supported by NSF EAGER Award \#2146691 and the ASU Center for Human, Artificial Intelligence, and Robot Teaming (CHART).}\thanks{All authors are with Arizona State University (ASU), Tempe, AZ 85287.}
\thanks{$^{a}$ S.~S.~Ulhas, A.~Ravichander, and S.~Berman are with the ASU School for Engineering of Matter, Transport and Energy~{\tt\small\{sulhas,aravic22,spring.berman\}@asu.edu}}
\thanks{$^{b}$ K.~A.~Johnson is with the ASU Department of Psychology~{\tt\small kathryn.a.johnson@asu.edu}}
\thanks{$^{c}$ T.~P.~Pavlic is with the ASU School of Computing and Augmented Intelligence, the School of Sustainability, and the School of Complex Adaptive Systems~{\tt\small tpavlic@asu.edu}}
\thanks{$^{d}$ L.~Gharavi is with the ASU School of Music, Dance and Theatre~{\tt\small lance.gharavi@asu.edu}}
}
\maketitle

\begin{abstract}
CHARTOPOLIS is a multi-faceted sociotechnical testbed meant to aid in building connections among engineers, psychologists, anthropologists, ethicists, and artists. Superficially, it is an urban autonomous-vehicle testbed that includes both a physical environment for small-scale robotic vehicles as well as a high-fidelity virtual replica that provides extra flexibility by way of computer simulation. However, both environments have been developed to allow for participatory simulation with human drivers as well. Each physical vehicle can be remotely operated by human drivers that have a driver-seat point of view that immerses 
them within the small-scale testbed, and those same drivers can also pilot high-fidelity models of those vehicles in a virtual replica of the environment. Juxtaposing human driving performance across these two contexts will help identify to what extent human driving behaviors are sensorimotor responses or involve psychological engagement with a system that has physical, not virtual, side effects and consequences. Furthermore, through collaboration with artists, we have designed the physical testbed to make tangible the reality that technological advancement causes the history of a city to fork into multiple, parallel timelines that take place within populations whose increasing isolation effectively creates multiple independent cities in one. Ultimately, CHARTOPOLIS is meant to challenge engineers to take a more holistic view when designing autonomous systems, while also enabling them to gather novel data that will  
assist them in making these systems more trustworthy.
\end{abstract}

\section{Introduction}

\begin{figure}
\centering
\includegraphics[width=\linewidth]{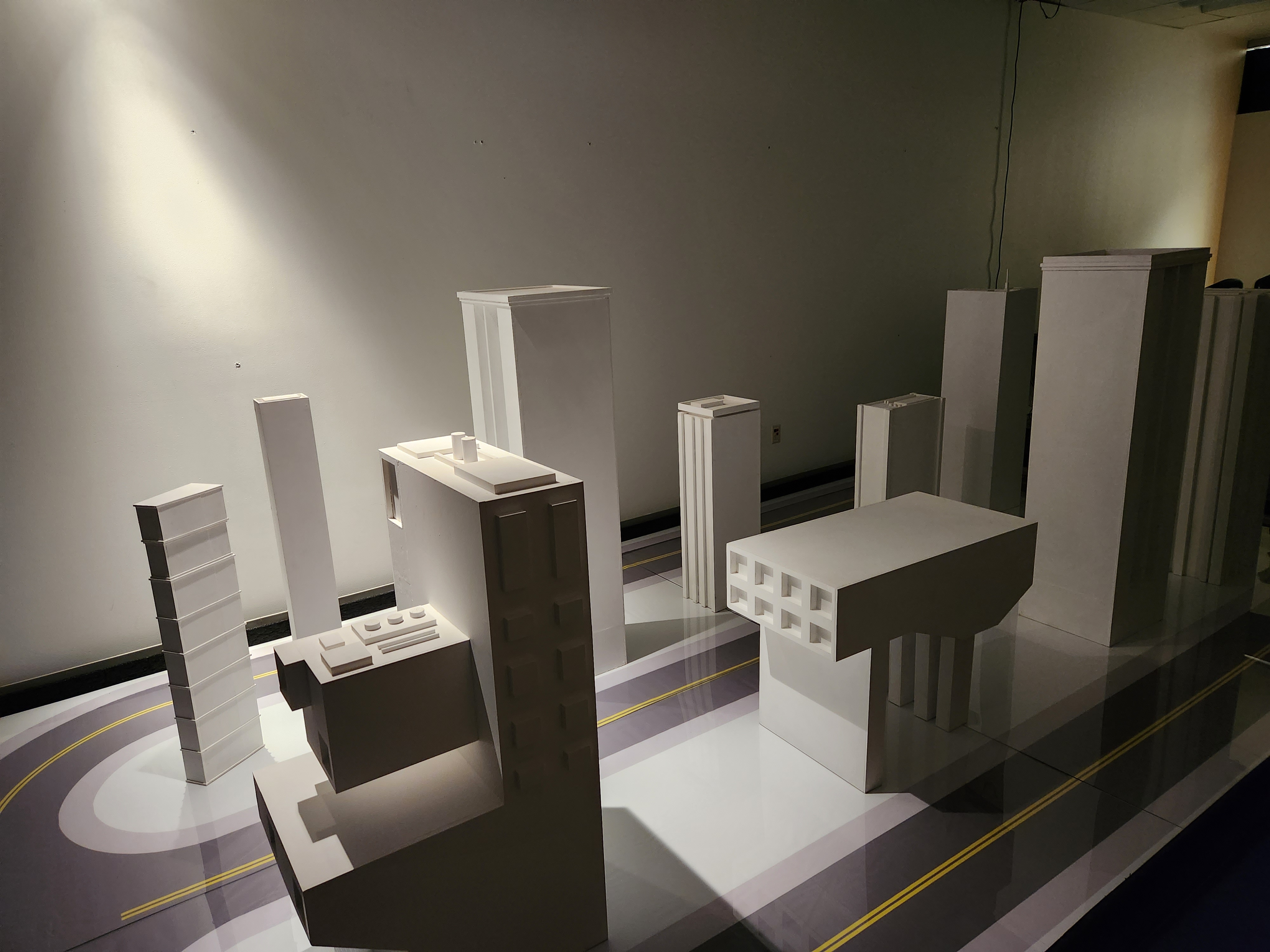}
\caption{CHARTOPOLIS, a reconfigurable, modular testbed for urban autonomous-vehicle research.}\label{fig:CHARTOPOLIS-2_image}
\end{figure}

We are developing \emph{CHARTOPOLIS}, a small-scale traffic testbed that serves as a research laboratory, an art installation, and a boundary object~\cite{SG89institutionalEcology} that is intended to increase integration between engineering, psychology, anthropology, and ethical philosophy. This mind--machine--motor nexus~(M3X)\footnote{NSF M3X program: \url{https://beta.nsf.gov/funding/opportunities/mind-machine-and-motor-nexus-m3x}} ``labor-art-ory'' consists of a model physical driving environment with small robotic vehicles~(Fig.~\ref{fig:CHARTOPOLIS-2_image}), a driving station for remote control of the vehicles by human participants~(Fig.~\ref{fig:DrivingStation}), and a driving simulator that serves as a high-fidelity, virtual replica of the physical environment~(Fig.~\ref{fig:DrivingSimulator}). 

In contrast to existing small-scale self-driving car testbeds, e.g.~\cite{Kloock2021, paull2017duckietown, stager2018scaled,hyldmar2019fleet,berger2014competition,vincke2021open, quanserstudio}, CHARTOPOLIS is specifically designed to facilitate participatory studies of human sensorimotor, behavioral, cognitive, and aesthetic responses to diverse driving scenarios with the goal of enriching autonomous vehicles with human-like behavioral profiles. Its matching virtual and physical environments will enable safe, controlled experimental manipulation of both typical driving conditions and unavoidable accidents that would be difficult or hazardous to replicate with full-scale vehicles. Furthermore, by comparing and contrasting human driving performance in the physical testbed to performance in the high-fidelity simulator, we can identify commonalities among human behaviors across the two participatory platforms~(i.e., the physical testbed and virtual replica) that are likely to extend to hypothetical behaviors in full-scale vehicles. Thus, this comparative approach aims to elucidate the underlying problems resulting in the Sim2Real gap rather than attempting to find costly and risk-prone stopgap solutions to them, e.g., \cite{huang2021autonomous, huang2021towards}. More generally, juxtaposing the physical and virtual environments enables us to investigate the minimal set of features~(e.g., sensory stimuli, dynamical characteristics, psychological association with outcomes in physical space) required for a participatory driving testbed to effectively engage a human operator in as realistic of a driving experience as possible. Finally, CHARTOPOLIS doubles as an art installation that makes a statement about the effect of technology and autonomy on the evolving history of a city. 

\section{Goals: Human-Focused Design and Art}

\subsection{Trustworthy Autonomy via Participatory CHARTOPOLIS}

Much of the focus in developing autonomous vehicles~(AVs) has been on improving sensors and algorithms to enable accurate perception and enhance driver safety. To that end, researchers and manufacturers have worked intensely at designing AVs that emulate human driving behavior, but little effort has been placed in determining \emph{which} humans to emulate. To what extent are we able to account for the significant variance in human drivers' personalities, temperaments, values, and moral priorities? Can we design AVs that reflect the personality types and driving styles of their owners? The CHARTOPOLIS testbed and driving simulator allow us to investigate this variability in human driving performance in a safe environment. 

\subsubsection{Mapping Parameters to Personality and Values}
Gaining the trust and acceptance of human passengers and human drivers of other vehicles will require fully autonomous vehicles to do more than be competent at obeying the objective rules of the road; AVs will also have to emulate human driving behaviors that are acceptable in terms of social and ethical norms. Within any group of human drivers who all obey driving laws, the remaining unconstrained degrees of freedom allow for significant differences to emerge across driving preferences~(e.g., following distance, responsiveness to light changes, responsiveness to upcoming speed-limit changes, etc.). These driving-style differences, in turn, reflect variability in individual drivers' motives, values, moral priorities, and other psychological states/traits. 

Consistent suites of different driving preferences can conspicuously identify a driver's ``personality'' as benevolent/careful/pessimistic/defensive or power-oriented/egoistic/optimistic/aggressive, and the resulting behaviors can be placed on a normative ethical scale. An AV's programmer is free to choose these driving parameters, possibly reflecting their own driving personality. In optimal control theory~\cite{dorfbishop2008modern}, such remaining degrees of freedom might map to some scalar functional~(e.g., energy use or some proxy for physical driving comfort) that can be optimized through an automated design process. In either case, no \emph{explicit} characterization of the \emph{ethical} dimension of these choices is incorporated into the design. A major motivation of CHARTOPOLIS is to develop a framework for formalizing the currently cryptic ethical dimension of sociotechnical systems' modeling and control design.

\subsubsection{Beyond the Artificial Moral Dilemma}
Prior attempts to characterize morality and ethical behavior in machine decision-making, e.g.~\cite{awad18moralmachine}, implicitly assume that ethical stances are only evident as the outcome of often contrived, singular, pathological driving events. For example, Awad et al.~\cite{awad18moralmachine} asked humans to judge hypothetical AV decision-making by using a battery of questions about driving-related dilemmas, such as whether an AV with a sudden braking failure should crash into a wall~(certainly killing its passengers) or continue driving into a crowded crosswalk~(certainly killing others on the road). Humans evaluating these two options had significantly more time to deliberate on the correct answer than the AV would have. Furthermore, some scenarios were only dilemmas in the myopic perspective; for example, crashing  into pedestrians in a crosswalk does not guarantee that the brake-less AV will not immediately afterward hit something else that will also kill everyone in the AV. Implicit in these studies is that the ideal AV will have an explicit rule-based (deontological) or utility-based (utilitarian) reasoning system, c.f.~\cite{gerdes2015implementable}, that will recognize emergent dilemmas and, at those instants, assert a (hopefully acceptable) decision.

In contrast with those prior attempts, we recognize that ethical stances are being made continually throughout the driving process. An ``aggressive'' driver might be viewed by an observer as behaving ``less ethically'' than a ``defensive'' driver despite neither of them actually being observed in a formal dilemma. These ethical stances emerge from the non-trivial combination of the human's driving preferences, sensing and actuation dynamics of the human and the vehicle, and the physical realities of the exterior world. In other words, ethical stances are an ecological property of the system of the driver's mind, the motor~(human sensing and actuation dynamics), and the machine~(physical dynamics of the vehicle and the surrounding environment). Engineers should characterize the ethical stances that their AVs implicitly take as they operate continually and also formally recognize how their technologies modulate the ethical stances taken by their human operators.

\subsection{CHARTOPOLIS as Art Installation}

CHARTOPOLIS will constitute a kind of multivalent work, not only in its function as laboratory but also in a dual function as artwork and as a kinetic installation whose collection of vehicles, buildings, roadways, signs, and inhabitants serve as a boundary object~\cite{SG89institutionalEcology} suggesting multiple interpretations. It will serve as a site for both the practice of science and for meditation on the worlds that science creates as well as a tool for thinking through technology, specifically robots and AI, as a dynamic between opposing imaginaries: salvation and damnation, utopia and dystopia, hope and dread. Robots, in the differences we draw between \emph{us} and \emph{them}, become a way to think about human values and the fantasy of automation as an ethically immaculate source of labor. Robots serve for us the function that Viktor Shklovsky~\cite{shklovsky1917art} identifies for art: they make us strange. 

Our ecological perspective of continually operating ethics in driving contexts acts on both short time scales and on very long, sociotechnical evolutionary time scales. The increased penetration of AVs in urban environments results in the continuous operation of sensing technologies that are aware of features familiar to human perception~(e.g., visible~light) as well as features that are totally invisible to humans~(e.g., electronic, subsonic, supersonic, etc.). As urban infrastructure is altered to better enable the performance of AVs, those in less AV-friendly areas of a city become hidden from those who begin to depend on AVs. Moroever, with physical separation comes cultural and historical separation.

CHARTOPOLIS will capitalize on the narrative ability of physical and computer simulations to illuminate these ethical dimensions of AV design and control, which unroll over a wide range of operational and evolutionary time scales. The testbed design draws on the utopian aesthetic of architectural models and their vision of a clean, hopeful future, as well as on the weird fiction of China Mi\'eville~\cite{mieville2015limits}. The completed testbed will embody two cities, one visible and one transparent, that represent two realities occupying the same space. The ``invisible city'' will evoke those things that are erased and made invisible, forgotten or ignored, left out of the fables of the past and visions of the future. 

\begin{figure}[t]
  \centering
  \includegraphics[width=0.95\linewidth]{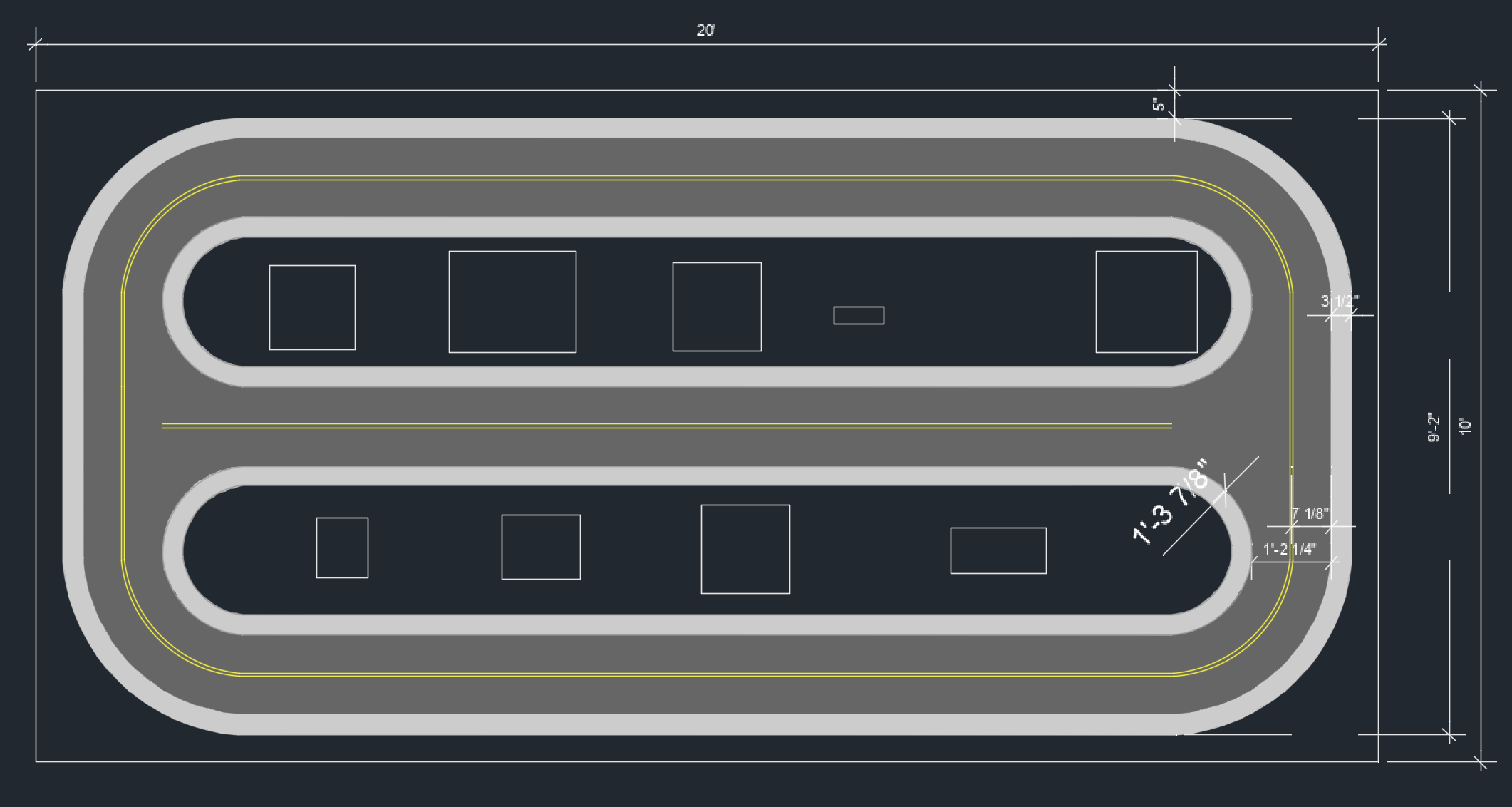}
  \caption{CHARTOPOLIS testbed layout with dimensions.}
  \label{fig:CHARTOPOLIS-layout}
\end{figure}

\begin{figure}[t]
  \centering
  \includegraphics[width=0.9\linewidth]{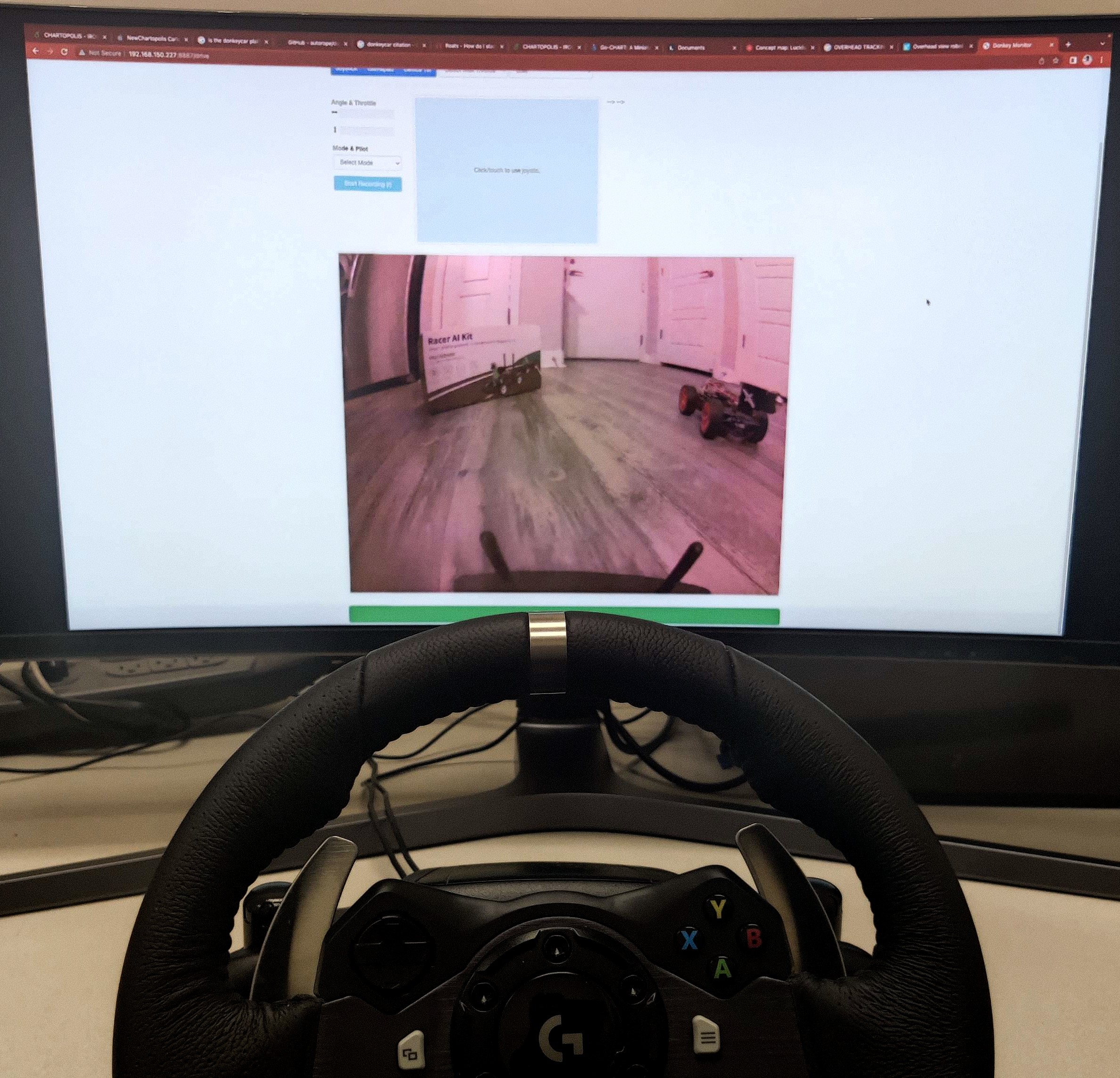}
  \caption{Computer station for remote operation of robotic car. Monitor displays a video stream from onboard camera.}
  \label{fig:DrivingStation}
\end{figure}

\begin{figure}[t]
  \centering
  \includegraphics[width=0.9\linewidth]{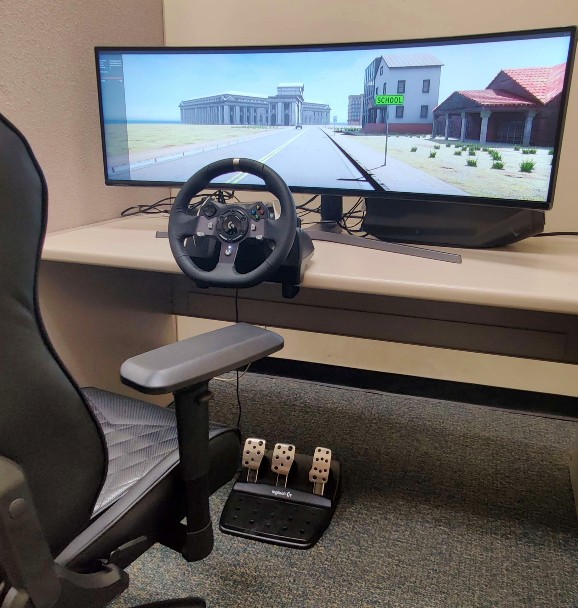}
  \caption{Driving simulator utilizing CARLA~\cite{dosovitskiy2017carla} environment.}
  \label{fig:DrivingSimulator}
\end{figure}

\section{Testbed Components}

\begin{figure}
  \centering
  \includegraphics[width=0.6\linewidth]{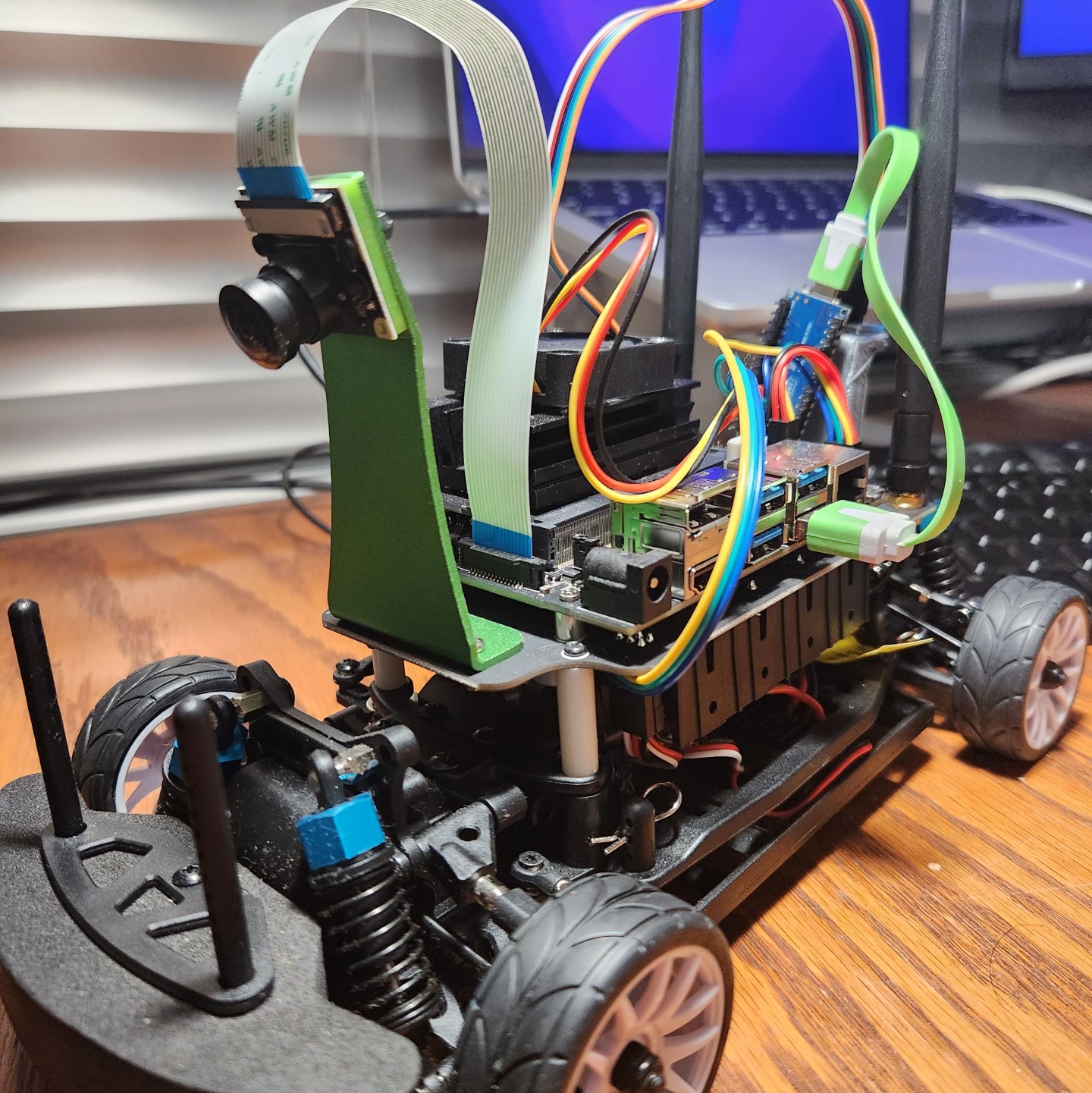}
  \caption{Modified JetRacer AI Pro robotic car~\cite{waveshare}.}
  \label{fig:JetRacer}
\end{figure}

We have developed several iterations of the CHARTOPOLIS testbed. The first~\cite{Subramanyam2018} consisted of a grid of roadways with traffic lights at intersections and several Pheeno~\cite{wilson2016pheeno} differential-drive robots, which were programmed with autonomous driving functions such as lane tracking and traffic light detection. The second version was created in conjunction with the Go-CHART~\cite{Kannapiran2020} miniature car robot, which emulates sensing and computation capabilities of a full-size AV, and included traffic lights, signs, and scenery (grass and trees). For this work, we also developed an initial version of our driving station for remote control of the robots. Our most recent version of CHARTOPOLIS~(Fig.~\ref{fig:CHARTOPOLIS-2_image}) enhances the versatility and portability of the testbed and the onboard compute capability of the robotic cars. This version comprises a customizable driving environment with roads, traffic lights and signs, reconfigurable buildings, and adjustable lighting, and it uses a modified version of the JetRacer Pro AI kit \cite{waveshare}~(Fig.~\ref{fig:JetRacer}) as the robot car platform. 

The following modifications were made to the Jetracer Pro AI Kit and its control interface. Photo-interrupt encoders were coupled with the four-wheel drivetrain's shaft to obtain feedback for speed control. An Arduino Nano was added to expand the I2C buses on the Jetson Nano and to facilitate the use of prebuilt Arduino libraries that are  not supported on the Jetson Nano. An IMU sensor, which uses the expanded I2C bus, was mounted on the back of the robot on an elevated Z-bracket in order to prevent damage to it from collisions. The Donkey Car web-control interface \cite{roscoe2019donkey} was modified to include the data from the encoders and IMU sensor and to improve the accuracy of remote steering control of the robot via the Logitech G920 steering wheel and pedals by mapping its throttle and steering angle to the PWM pulses sent from the PCA9685 I2C controller board.

In both the remote-control driving station (Fig. \ref{fig:DrivingStation}) and driving simulator (Fig. \ref{fig:DrivingSimulator}), a human operator has a first-person-view of the roadway on a monitor and drives the simulated or physical car using a Logitech G920 steering wheel with accelerator and brake pedals. We are modeling different driving scenarios in the driving simulator using the open-source virtual driving environment CARLA \cite{dosovitskiy2017carla}. This builds on our previous work developing a driving simulator \cite{Ulhas2019,Kankam2019} to obtain data for our earlier set of studies on human driving responses (unpublished), described in Section \ref{sec:exp}. The environment in our simulations is an exact replica of the road layout (Fig.~\ref{fig:CHARTOPOLIS-layout}) and buildings on the CHARTOPOLIS testbed. The layout is replicated to scale in OpenDRIVE format using RoadRunner \cite{Roadrunner}, and the buildings are imported as assets through the Unreal Engine editor. The complete simulated CHARTOPOLIS is finally packaged as a portable CARLA distribution.

\begin{figure}[t]
  \centering
  \includegraphics[width=.95\linewidth]{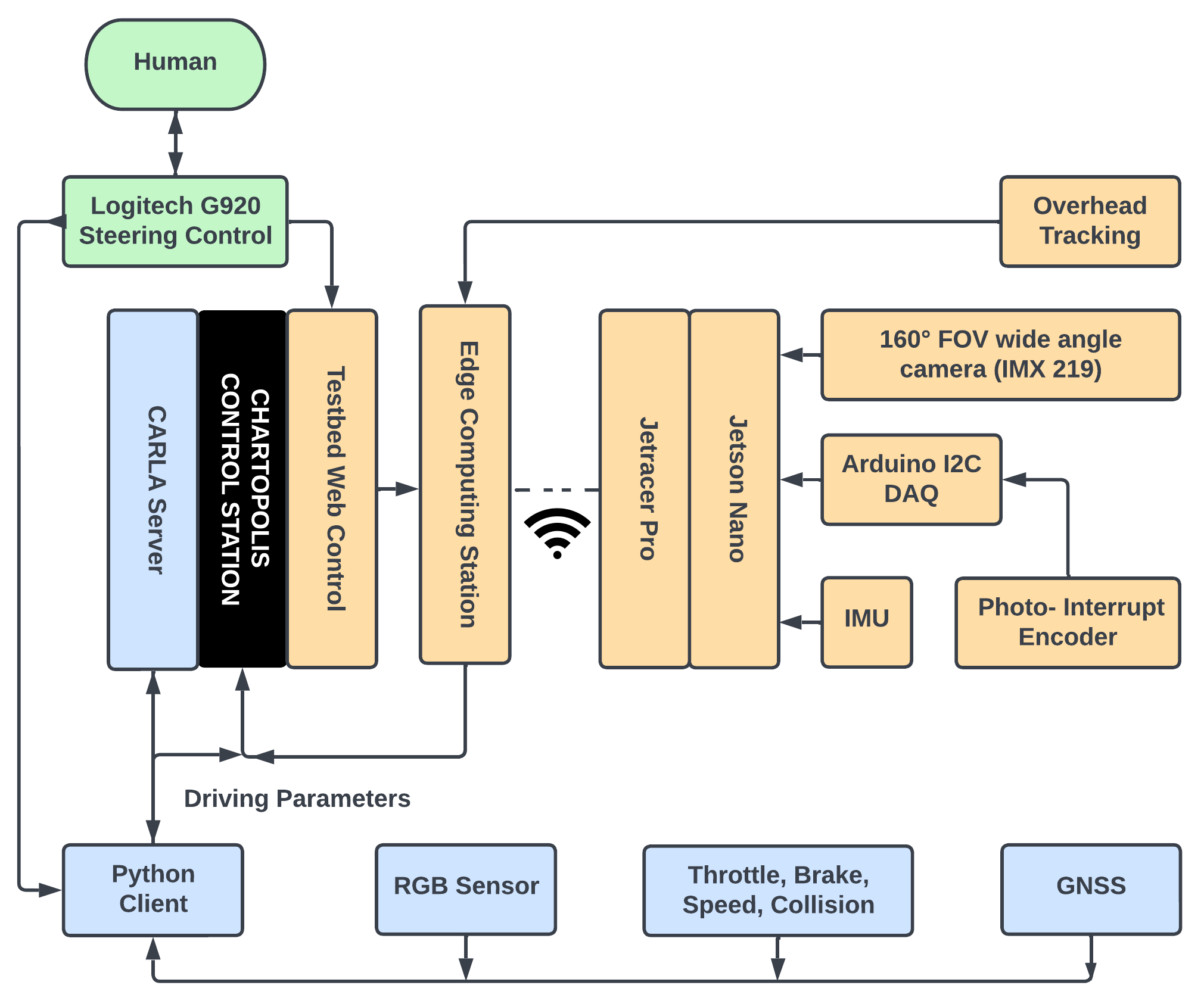}
  \caption{CHARTOPOLIS control architecture, including components of the human-robot interface ({\it green}), physical testbed ({\it orange}), and CARLA driving simulator ({\it blue}).}
  \label{fig:ControlArchitecture}
\end{figure}

The control architecture of the CHARTOPOLIS testbed and simulator is illustrated in Fig.~\ref{fig:ControlArchitecture}. A single interface common to the physical and simulated environments helps to characterize and mitigate the Sim2Real gap by allowing for direct comparison of performance across the two environments. In the physical testbed, the robot's pose in a global coordinate frame is obtained using an overhead camera; further experiments will determine whether a motion-capture system is necessary to achieve a sufficiently accurate mapping of the robot's pose between the physical testbed and simulation. The robot measures its velocity using its onboard encoder and obtains image data from its wide-angle camera. The pose, velocity, and images are broadcast to the Jetson Nano, which computes the robot's control actions using this feedback. In the simulator, the CARLA Server obtains the virtual vehicle's state from simulated GNSS and images from its onboard RBG sensor, and the Python Client uses these data to compute the vehicle's control actions.

\section{Experimental Procedures} \label{sec:exp}

In our planned studies, human participants will be given a battery of questionnaires including: personality traits~\cite{john1999big}, values~\cite{schwartz2012refining}, moral priorities~\cite{graham2011mapping}, and driving style~(e.g., positive vs.~aberrant)~\cite{ozkan2005new}. These same participants will then be invited to the lab, where they will experience driving in both the simulator and through remote control of a robot car within the physical testbed with a first-person video livestream (counterbalanced for ordering effects).
	
Our data-analysis plan will be to map the pre-screened personality profiles and self-reported driving styles of the human drivers onto the driver behavior data collected in the virtual and physical environments. Our previous, unpublished data show that individuals with power-oriented vs. benevolent profiles are more likely to have self-reported aggressive vs. positive (prosocial) driving styles, respectively. Further, the two driving styles are highly predictive of the number of real traffic violations (power-oriented, aggressive drivers having significantly more violations). We expect these traits and driving styles to be evident in the driving behaviors in the simulator and matching physical testbed. Ultimately, this information will allow us to design controllers that mimic, at least these two, personality and driving styles.

Gathered simulation data will include all state-variable and sensor data that are necessary to reproduce aspects of the driving experience, including: the vehicle's ground-truth position, speed, and acceleration; data from the onboard navigational sensors (camera, IMU, GNSS); and information about lane changes and collisions. On the physical robot, speed and acceleration data will be calculated using sensor fusion of data from a photo-interrupt shaft encoder on the robotic drivetrain with IMU and positional data from an overhead camera, and these data will be synchronously recorded. In both the simulator and testbed, we will collect data on steering, pedal angle, and braking.

\section{Conclusion and Open Challenges}

The CHARTOPOLIS labor-art-ory is a work in progress, with further technical and conceptual challenges to overcome. The many challenges of implementing ethics in AVs are beyond the scope of this paper, but they warrant further consideration~\cite{bigman2018people, gerdes2015implementable, lin2016ethics}. One take-away message, however, is that all stakeholders should be consulted in implementing machine ethics~\cite{stilgoe2020developing}.

\section{Acknowledgement}
The authors thank Brunella Provvidente for helping with the design and implementation of the physical testbed.

\bibliographystyle{./bibliography/IEEEtran}
\bibliography{./bibliography/IEEEabrv,CHARTOPOLISReferences.bib}

\end{document}